\documentclass{article}

\usepackage{microtype}
\usepackage{graphicx}
\usepackage{booktabs} 

\usepackage[T1]{fontenc}
\usepackage[polish, english]{babel}
\usepackage[utf8]{inputenc}

\usepackage[accepted]{icml2024/icml2024}

\usepackage{amsmath}
\usepackage{amssymb}
\usepackage{mathtools}
\usepackage{amsthm}

\theoremstyle{plain}

\theoremstyle{definition}

\theoremstyle{remark}

\usepackage[textsize=tiny]{todonotes}

\usepackage[switch]{lineno}
\usepackage{subcaption}
\usepackage{float}
\usepackage{multirow}
\usepackage{soul}

\def\x{ {\bf x} }

\def\F{\mathcal{F}}
\def\L{\mathcal{L}}

\def\R{\mathbb{R}}

\def\our{HyperPlanes}

\usepackage[T1]{fontenc}
\usepackage[utf8]{inputenc}
\usepackage[polish]{babel}

\usepackage{hyperref}
\usepackage[capitalize,noabbrev]{cleveref}
\begin{document}

\twocolumn[
\icmltitle{\our{}: Hypernetwork Approach to Rapid NeRF Adaptation}



\icmlsetsymbol{equal}{*}
\begin{icmlauthorlist}
\icmlauthor{Paweł Batorski}{equal,ju,sd}
\icmlauthor{Dawid Malarz}{equal,ju}
\icmlauthor{Marcin Przewięźlikowski}{ju,sd,ideas}
\icmlauthor{ Marcin Mazur}{ju}
\icmlauthor{ Slawomir Tadeja}{st}
\icmlauthor{Przemysław Spurek}{ju}
\end{icmlauthorlist}

\icmlaffiliation{ju}{Jagiellonian University, Faculty of Mathematics and Computer Science, Cracow, Poland
}
\icmlaffiliation{sd}{Jagiellonian University, Doctoral School of Exact and Natural Sciences, Cracow, Poland}
\icmlaffiliation{ideas}{IDEAS NCBR}
\icmlaffiliation{st}{Department of Engineering, University of Cambridge, Cambridge, UK}

\icmlcorrespondingauthor{Paweł Batorski}{pawel.batorski@doctoral.uj.edu.pl}

\icmlkeywords{Machine Learning, ICML}

\vskip 0.3in
]



\printAffiliationsAndNotice{\icmlEqualContribution} 

\begin{abstract}

Neural radiance fields (NeRFs) are a widely accepted standard for synthesizing new 3D object views from a small number of base images. However, NeRFs have limited generalization properties, which means that we need to use significant computational resources to train individual architectures for each item we want to represent. 
To address this issue, we propose a few-shot learning approach based on the hypernetwork paradigm that does not require gradient optimization during inference. The hypernetwork gathers information from the training data and generates an update for universal weights.
As a result, we have developed an efficient method for generating a high-quality 3D object representation from a small number of images in a single step. This has been confirmed by direct comparison with the state-of-the-art solutions and a comprehensive ablation study.

\end{abstract}

\section{Introduction}
\label{sec:intro}


Neural Radiance Field (NeRF)~\cite{mildenhall2020nerf} is a fully connected, non-convoluted neural network architecture that can generate new, vivid renderings of complex 3D objects from a handful of images. The latter has to be taken from known viewpoints as NeRF represents a given object using a 5D vector comprised of the coordinates of the camera's position and viewing direction. 
As we render the coloring and density along rays passing through the scene, the object's shape and color scheme are directly encoded in the neural network's weights with the loss function inspired by classical volume rendering~\cite{kajiya1984ray}. Consequently, NeRF can synthesize high-quality views of 3D objects utilizing the relationship between the small set of input images and well-known computer graphics methods (e.g. ray tracing or radiosity)~\cite{mildenhall2020nerf}. 

These unique characteristics make NeRF an ideal candidate to help tackle shortcomings experienced by other emerging technologies that classical computer graphics cannot fully resolve \cite{9830072}. For example, the seamless generation of plausible 3D objects is a crucial, underexplored challenge, especially in the case of content generation for computer games \cite{9830072}, virtual training \cite{Bozzi3DPrinter}, and other application areas relying on spatial graphical interfaces, e.g., virtual reality (VR) \cite{MedicalVR} or augmented reality (AR) \cite{Bozzi3DPrinter}. Here, NeRF offers close-to-revolutionary advantages over the classical methods, such as their speed, quality of renders, or the ability to generate complex 3D objects from a handful of images~\cite{mildenhall2020nerf} that populate such virtual environments \cite{NeRFVR}. 

\begin{figure}
    \centering

    \begin{tikzpicture}[scale=0.4]
    \node[inner sep=0pt] (russell) at (-5.0,0)
    {\includegraphics[trim={0cm, 0cm, 0,3cm, 0cm},clip,width=\linewidth]{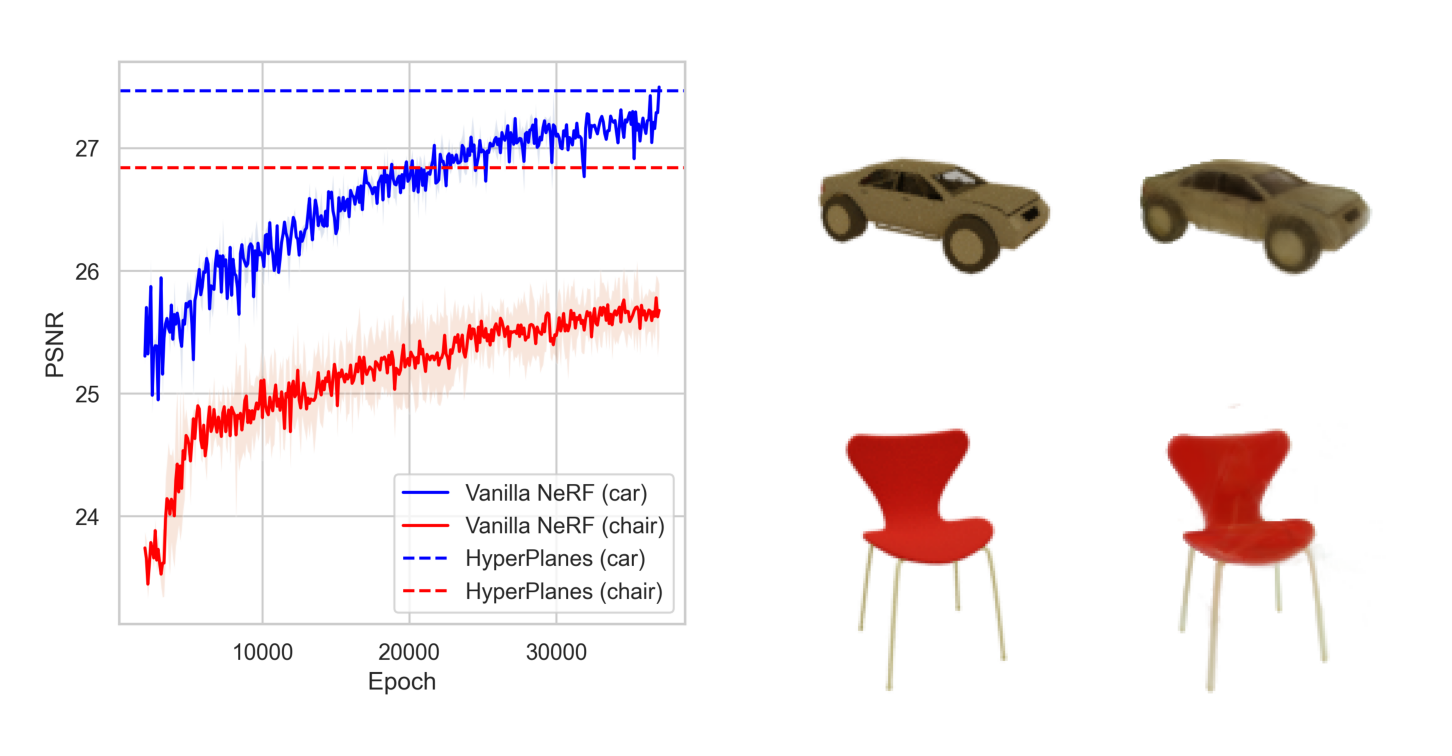} };
    \node[text width=3.5cm] at (0.0,4.2) {\footnotesize Ground truth };
    \node[text width=3.5cm] at (5.0,4.2) {\footnotesize \our{} };
    \end{tikzpicture}

    \caption{Our \our{} model can ({\em in a single update}) produce 3D objects that achieve a superior PSNR ($\uparrow$) when compared to those generated by a vanilla NeRF trained for approximately 36000 epochs. 
    This results in \emph{$380 \times speedup$} (10 seconds for adapting and rendering with \our{} vs. 63.7 minutes for training NeRF) in object reconstruction speed while achieving better or comparable quality.
    } 
    \label{fig:single_update_vs_nerf}
\end{figure}

However, NeRF also suffers from significant constraints. Notably, it has to be trained separately on each object as the architecture does not generalize to new, previously unencountered data \cite{liu2020neural}. In addition, training times are often lengthy since the neural network's weights must encode the object's shape characteristics \cite{liu2020neural}.

To overcome these constraints, we can employ classical generative models to force NeRF to generalize to unseen objects. Here, the existing literature provides examples of GANs (e.g., GRAF~\cite{schwarz2020graf}, $\pi$-GAN~\cite{chan2021pi}, EG3D \cite{chan2022efficient}), autoencoders (e.g., Point2NeRF~\cite{zimny2022points2NeRF}), and diffusion models (e.g., NFD~\cite{shue20233d}, NeRFDiff~\cite{gu2023NeRFdiff}, SSDNeRF~\cite{chen2023single}). Such models can generate new 3D objects, but using them to produce NeRF representations from existing images is complex and typically requires lengthy, resource-consuming, and costly training \cite{zimny2022points2NeRF}. In contrast, to make it practical and applicable in real-life scenarios, we want to take just a few photographs with a smartphone or any other mass-market video-capturing device and convert them to NeRF without unnecessary training \cite{mueller2022instant}.

To address the mentioned problem, we can use meta-learning techniques such as MAML~\cite{finn2017model} or one of its extensions (e.g., Reptile~\cite{nichol2018first}). The main idea is to find such parameters of a model that can be easily adapted to a new task in a few or even a single gradient step. Thus, MAML algorithms aim to find model weights sensitive to task changes. Small parameter shifts will significantly improve the loss function of any task drawn from the tasks' distribution when changed in the direction of its gradient. Unfortunately, using such approaches to fit a high-quality NeRF model needs a few hundred gradient steps to update the universal weights. Moreover, a few thousand updates are required in inference time \cite{tancik2021learned}.

In contrast to all the presented methods and their inherent limitations, we propose to rely on the hypernetwork paradigm~\cite{sendera2023hypershot, przewikezlikowski2022hypermaml} for a few-shot training. Such an approach does not require gradient optimization in the inference time. Instead, hypernetwork aggregates information from training data and produces an update for universal weights. Thus, by combining hypernetworks and NeRFs with partially non-trainable MultiPlaneNeRFs \cite{zimny2023multiplanenerf}, we have developed an efficient method for generating 3D representations from 2D images in a single step. To that end, we contribute in this paper a new method called~\our{} that uses the hypernetworks in NeRF training. The resulting model can generate a NeRF representation from a few existing images in a single step. Consequently, it can instantly generate views of new, unseen objects, as demonstrated through extensive experimental studies presented in our paper. 

\section{Related works}
\label{gen_inst}


There are many ways in which we can represent 3D objects, including point clouds \cite{yang2022continuous}, octrees \cite{hane2017hierarchical}, voxel grids~\cite{choy20163d}, multi-view images \cite{LIU2022108774}, or deformable meshes \cite{li2017learning}. In contrast to these discrete representations, NeRF~\cite{mildenhall2020nerf} relies on a fully connected, non-convoluted neural network to store information about the scene. Thus allowing the rendering of new, formerly unseen views of a given 3D object \cite{mildenhall2020nerf}.

\paragraph{Representations and rendering}
To generate new views of a given scene, the NeRF model or any of its multiple generalizations uses differentiable volumetric rendering  \cite{liu2020neural, barron2021mip, niemeyer2022regNeRF, roessle2022dense}. These architectures, however, have their own caveats, such as extensive training time. 

    \begin{figure*}
    \begin{subfigure}[t]{0.48\textwidth}
        \centering
        \includegraphics[width=\textwidth]{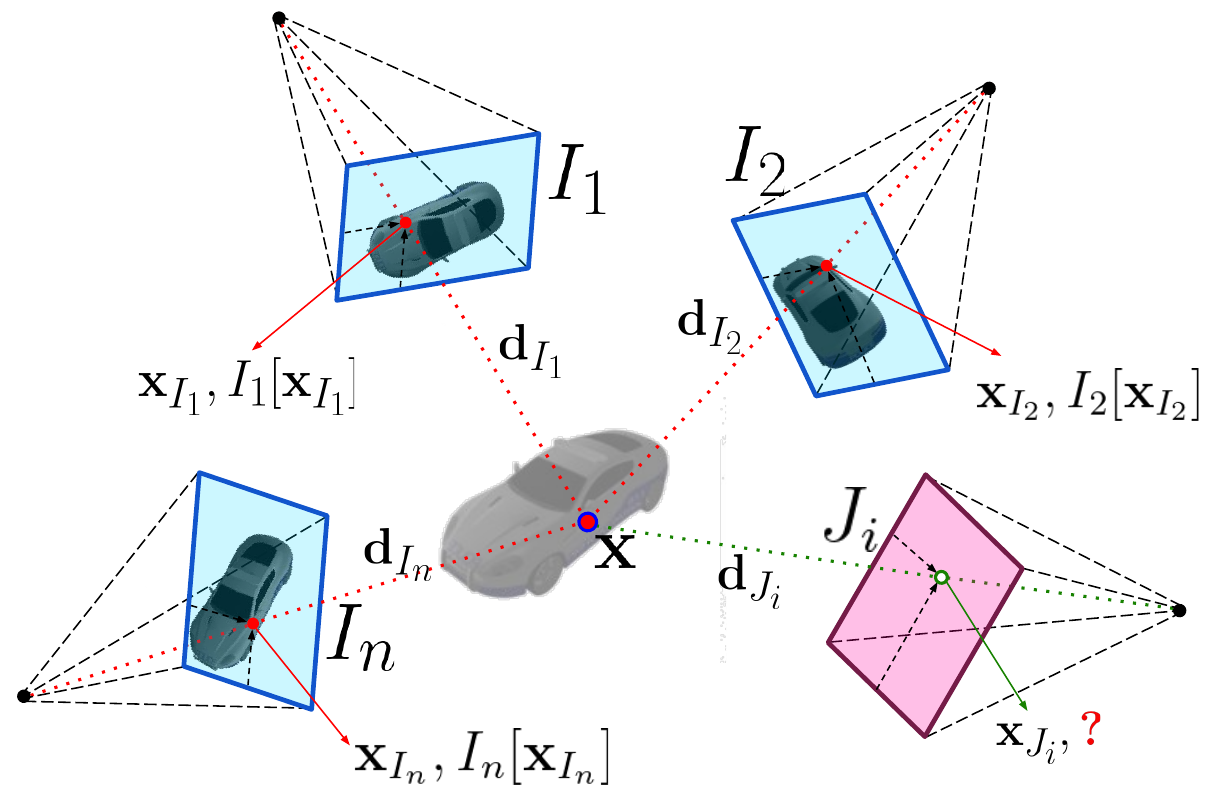}
        \caption{Data processed by the \our{} model. We consider support ImagePlanes $I_1, \ldots, I_n$ with corresponding viewing directions $\mathbf{d}_{I_1},\ldots,\mathbf{d}_{I_n}$. 
        Given a point $\mathbf{x}$, we project it onto $I_1, \ldots, I_n$, obtaining the points $\mathbf{x}_{I_1},\ldots, \mathbf{x}_{I_n}$, and interpolate their colors $I_1[\mathbf{x}_{I_1}],\ldots, I_n[\mathbf{x}_{I_n}]$. The goal is to render the color of the projection of $\mathbf{x}$ onto each query ImagePlane $J_i$ (related to a viewing direction $\mathbf{d}_{J_i}$). 
        }
        \label{fig:setting}
    \end{subfigure}
    \hfill
    \begin{subfigure}[t]{0.48\textwidth}
        \centering
    \includegraphics[width=\textwidth]{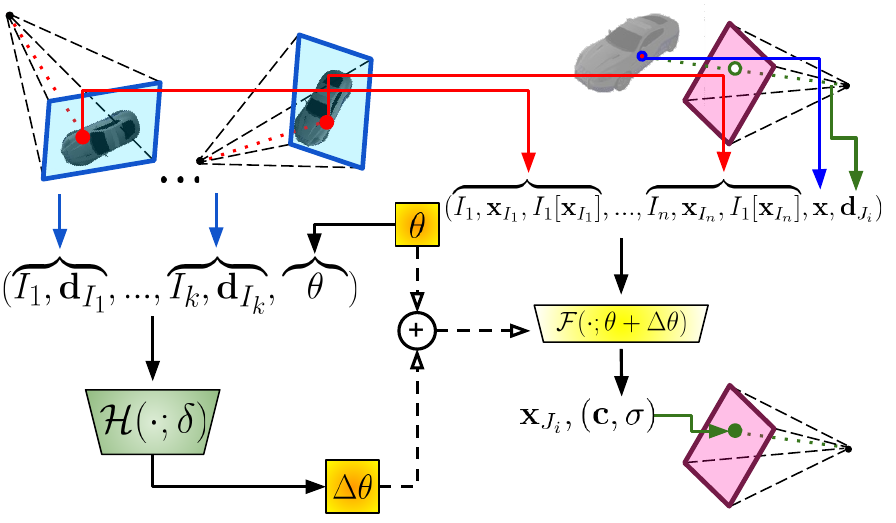}
        \caption{Architecture of the \our{} model. Hypernetwork $\mathcal{H}$ predicts updates $\Delta \theta$ to parameters $\theta$ of target network $\mathcal{F}$ (PointMultiPlaneNeRF) based on selected support ImagePlanes $I_1, \ldots, I_k$, their respective viewing directions $\mathbf{d}_{I_1},\ldots, \mathbf{d}_{I_k}$, and $\theta$ itself. Next, $\mathcal{F}$ (now parameterized with $\theta + \Delta \theta$) infers the color $\mathbf{c}$ and the volume $\sigma$ of the projection $\mathbf{x}_{J_i}$ of the point $\mathbf{x}$ onto a query ImagePlane $J_i$ (related to a viewing direction $\mathbf{d}_{J_i}$) using Eq.~\eqref{eq:pointmpnerf}.}
    \end{subfigure}
    \caption{Visualization of data utilized by \our{} (left), and architecture (right). }
        \label{fig:schema}
\end{figure*}
To address these limitations, numerous new approaches have been proposed. For instance, a 3D asset can be represented by an orthogonal tensor component \cite{chen2022tensorf}, or we can extend the NeRF model with point clouds or depth maps \cite{wei2021NeRFingmvs, deng2022depth, roessle2022dense}.

Another set of methods focused on voxel grids. We can optimize such feature voxel grids to warrant fast radiance field reconstruction \cite{sun2022direct}, which can also be split into a separate, multilevel grid \cite{muller2022instant}. Moreover, a sparse voxel grid can use its nodes as repositories with spherical harmonic coefficients and densities \cite{fridovich2022plenoxels}.


Another group of NeRF models are built to use only a handful of input images \cite{chen2021mvsNeRF, yu2021pixelNeRF, wang2021ibrnet}. The majority of these techniques utilize a form of a feature extractor generalizable thanks to training on a diversified catalogue of objects  \cite{wang2021ibrnet}. Furthermore, 3D CNN can be employed to aggregate information from multiple sweeping planes warping 2D neural features. The assemblage process is then carried out by a ray transformer network \cite{chen2021mvsNeRF}. Other authors decided to add a convolution layer to transfer input views as means of representing the NeRF model \cite{yu2021pixelNeRF}.

An alternative approach to produce generalized NeRF is to use generative models. For example, HoloGAN~\cite{nguyen2019hologan} and BlockGAN~\cite{nguyen2020blockgan} use an implicit 3D representation to model real-world objects.

In GRAF~\cite{schwarz2020graf} and $\pi$-GAN~\cite{chan2021pi}, the authors use implicit neural radiance fields to generate 3D-aware images and geometry. In ShadeGAN~\cite{pan2021shading}, the shading-guided pipeline is used, and in GOF~\cite{xu2021generative}, the sampling area of each camera ray is gradually reduced.
In GIRAFFE~\cite{niemeyer2021giraffe}, low-resolution feature maps are generated first. Then, a 2D CNN is used to produce higher resolution outputs.

In StyleSDF~\cite{or2022stylesdf}, the authors merge an SDF-based 3D representation with a StyleGAN2 for image generation. In EG3D~\cite{chan2022efficient}, the StyleGAN2 generator and a TriPlane representation of 3D objects are utilized. Such models outperform other solutions in the quality of generated objects but are extremely difficult to train~\cite{chan2022efficient}.

The TriPlane representation from EG3D~\cite{chan2022efficient} finds many applications in diffusion-based generative models. In \cite{shue20233d}, the authors propose to train the diffusion model to generate a TriPlane representation. The model is trained in two steps. First, we replace the data by a TriPlane representation, after which, we train the final diffusion model. A similar strategy is used in \cite{muller2023diffrf}, but with the voxel representation instead of the TriPlane. In \cite{gu2023NeRFdiff}, the authors propose to modify the U-Net component in diffusion models. On the other hand, \cite{ssdNeRF} introduces a single-step training for the NeRF-based diffusion model.

In \cite{zimny2023multiplanenerf}, the authors propose MultiPlaneNeRF, which is an alternative model to the TriPlane decoder that uses existing images as a non-trainable representation. In our project, we use the MultiPlaneNeRF architecture because we do not need a large feature extractor, which is essential in hypernetwork-based few-shot learning.

\paragraph{Few-shot learning}

One of the most crucial limitations of contemporary deep learning models is the requirement for large amounts of data to train them~\cite{wang2020fewshotsurvey}. 
To address this, a wealth of few-shot learning (FSL) approaches have been proposed in recent years, including ~\cite{munkhdalai2017meta, munkhdalai2018rapid, raghu2019rapid, ye2020few, zhmoginov2022hypertransformer} and others\footnote{For a comprehensive survey of different few-shot learning techniques, see \cite{wang2020fewshotsurvey}.}. 
Typically, such methods involve training in a meta-learning framework, wherein they learn to solve disjoint diverse tasks, each defined by a support set -- a small amount (i.e. few shots) of labeled data, sampled from a large dataset~\cite{wang2020fewshotsurvey,hospedales2020metalearning}.  

The most popular family of FSL approaches are the optimization-based methods, which train general models that can adapt their parameters to the support set at hand in several gradient steps~\cite{finn2017model,nichol2018first,raghu2019rapid,rajeswaran2019meta,finn2018probabilistic}.
While the models such as MAML~\cite{finn2017model} and Reptile~\cite{nichol2018first} have been developed primarily for classification tasks, they have also been successfully used for adapting NeRF representations for rendering new objects~\cite{tancik2021learned}. 
Nevertheless, to obtain sharp renderings in practice, the model must be updated for a few thousand gradient updates, which increases its inference time and computational footprint, limiting its usefulness in practice.

Also, hypernetworks~\cite{ha2016hypernetworks} have been previously proposed to alleviate the mentioned problem of few-shot classification models by producing an update in one step without gradient optimization in inference time~\cite{sendera2023hypershot,przewikezlikowski2022hypermaml}. In contrast, in this paper we present a hypernetwork that achieves this goal for NeRF representations.

\section{\our{}: hypernetwork approach to one-shot NeRF adaptation}

This section introduces our novel \our{} model. Specifically, we first present the background and notation for NeRF architecture, training, and adaptation. Next, we describe a method for using the hypernetwork paradigm to construct a NeRF representation from 2D images without needing gradient optimization during inference.

\paragraph{Neural Radiance Field (NeRF)}

Vanilla NeRF~\cite{mildenhall2020nerf} is a model for representing individual complex 3D scenes using a neural architecture. It takes a 5D vector as input comprised of spatial coordinates $\x = (x, y, z)$ and a viewing direction ${\bf d} = (\phi, \psi)$. It uses these data to learn rendering the emitted RGB color ${\bf c} = {\bf c}({\bf x},{\bf d})$ and the volume density $\sigma=\sigma({\bf x})$. Roughly speaking, NeRF training boils down to minimizing a total squared error between the expected color $\hat{C}({\bf r})$ of a camera ray ${\bf r}={\bf o}+t{\bf d}$ with near and far scene bounds $t_n$ and $t_f$ (conceptually expressed as an integral over $t\in [t_n,t_f]$ involving the volume density $\sigma({\bf r}(t))$ and the color ${\bf c}({\bf r}(t),{\bf d})$, but numerically estimated by means of stratified sampling) and its true color $C({\bf r})$. Formally, the goal is to optimize the multilayer perceptron (MLP) network
\begin{equation}
\begin{array}{l}
\mathcal{F}_\text{NeRF} (\x , {\bf d}; \theta ) = ( {\bf c} , \sigma)
\end{array}
\end{equation}
by applying the following loss function:
\begin{equation}
\begin{array}{l}
    \mathcal{L}_{\text{NeRF}}(\cdot;\theta) = \sum _{{\bf r}} \| \hat C({\bf r}) - C({\bf r}) \|_2^2,
    \label{eq:cost_general}
\end{array}
\end{equation}
where the sum is taken over the set of sampled rays stored in batches and passing through the scene.

In practice, NeRF encodes the structure of an individual 3D object in neural network weights. Such an approach has important downsides, including time-consuming training needed separately for each 3D object and weak generalization to unseen data. A possible answer to these problems is using a MultiPlaneNeRF architecture, which we briefly outline in the next paragraph.

\paragraph{MultiPlaneNeRF}

The difference between NeRF and MultiPlaneNeRF \cite{zimny2023multiplanenerf} architectures is that the latter uses pre-existing 2D images, which are a non-trainable planar representation of a 3D object, to train a lightweight implicit decoder that aggregates image-based input to produce NeRF output. More formally, given a set $\mathcal{I}=\{I_1, \ldots, I_n\}$ of fixed 2D images, referred to as ImagePlanes, representing different views of a considered 3D scene, for any point ${\bf x} = (x,y,z)\in \R^3$ we first project it onto each $I_i$, obtaining a collection of 2D positions ${\bf x}_{I_1}, \ldots, {\bf x}_{I_n}$, 
and then 
linearly interpolate the RGB color $I_i[{\bf x}_{I_i}]\in \R^3$ of the point ${\bf x}_{I_i}\in \R^2$ from the colors of its four nearest pixels.
As a result, we have an input ${\bf z}={\bf z}({\bf x},\mathcal{I})\in \R^{5n}$ (consisting of all ${\bf x}_{I_i}$s and $I_i[{\bf x}_{I_i}]$s) to an implicit decoder that returns the color ${\bf c}={\bf c}({\bf z}, {\bf d})$ and the volume density $\sigma=\sigma({\bf z})$ using the NeRF rendering procedure. The resulting small MLP network
\begin{equation}
\begin{array}{l}
\F_\text{MultiPlaneNeRF} ({\bf z}, {\bf d}, \mathcal{I}; \theta) = ( {\bf c}, \sigma)
    \label{eq:implicitdecoder}
\end{array}
\end{equation}
is then trained with the loss function $\mathcal{L}_{\text{MultiPlaneNeRF}}(\cdot,\mathcal{I};\theta)$ as in Eq.~\eqref{eq:cost_general}.

As the authors of \cite{zimny2023multiplanenerf} show, MultiPlaneNeRF is a computationally efficient 3D scene representation model with good generalization capabilities when trained on a large dataset. However, it still suffers from lower rendering quality than vanilla NeRF, which has the advantage of learning each object separately. One of the methods to overcome this problem is to apply an appropriate few-shot parameter adaptation. Before discussing the details, it is important to note that our further studies are based on a modified version of the MultiPlaneNerf architecture. Namely, we assume that the color-volume outcome depends on both the point ${\bf x}$ and its 2D projections ${\bf z}$. This leads to the following neural model, referred to as PointMultiPlaneNeRF:
\begin{equation}\label{eq:pointmpnerf}
\begin{array}{l}
\F_\text{PointMultiPlaneNeRF} ({\bf x}, {\bf z}, {\bf d}, \mathcal{I}; \theta) = ( {\bf c}, \sigma),
\end{array}
\end{equation}
where ${\bf c}={\bf c}({\bf x},{\bf z},{\bf d})$ and $\sigma=\sigma({\bf x}, {\bf z})$.

\paragraph{Few-shot NeRF adaptation}

Brain-inspired few-shot learning methods aim to train a model that can easily adapt to unseen data using only a few samples or training iterations. In a simple meta-learning scenario used in our approach \cite{wang2020fewshotsurvey}, we deal with a series of randomly selected tasks $\mathcal{T}$, each consisting of a corresponding loss $\mathcal{L}$ and data distribution $q$ (from which we draw all samples). During training, the model learns a new task $\mathcal{T}$, typically using several gradient descent iterations for $\mathcal{L}$, from a few support samples. Next, it is tested on different query samples. The model update depends on the influence of parameter changes on the test error. It is adapted to a new task during inference by applying the same training procedure with support samples starting from the learned optimal parameters.

When applying a few-shot learning approach to a PointMultiPlaneNeRF architecture, the task is related to a representation of a single 3D object and consists of the corresponding loss expressed in Eq.~\eqref{eq:cost_general} and ImagePlanes retrieved from a dataset of its existing 2D views.
However, if we use a common implementation (e.g. MAML\footnote{Note that to update the weights MAML also uses time-consuming ``gradient through gradient'' second-order computations.}), many gradient descent iterations may be required, especially during inference, to ensure superior rendering performance. To avoid this problem, our proposed solution incorporates the hypernetwork paradigm, briefly outlined in the following paragraph.

\paragraph{Hypernetworks}

Hypernetworks \cite{ha2016hypernetworks} are neural models that generate weights for a target network to solve a specific task. In our \our{} approach, we take inspiration from recent successes of hypernetworks in few-shot learning~\cite{sendera2023hypershot,przewikezlikowski2022hypermaml} within the NeRF adaptation scenario. Here, such techniques reduced the number of iterations on support samples to a single step without gradient optimization. 

\paragraph{\our{}}

To formally summarize the above discussion, consider a few-shot learning scenario with individual tasks of the form
\begin{equation}\label{eq:nerftask}
\begin{array}{l}
    \mathcal{T}=\{\mathcal{L}_{\text{PointMultiPlaneNeRF}}(\cdot,\mathcal{S};\theta),
    \mathcal{S},\mathcal{Q}\},
\end{array}
\end{equation}
where $\mathcal{L}_{\text{PointMultiPlaneNeRF}}(\cdot,\mathcal{S};\theta)$ is the loss expressed by Eq.~\eqref{eq:cost_general} (replacing NeRF with the corresponding PointMultiplaneNeRF), while $S = \{I_1,...,I_n\}$ and $\mathcal{Q}=\{J_1,\ldots,J_n\}$ denote disjoint sets of support and query ImagePlanes retrieved from a dataset of existing 2D views of a given 3D object.
Our goal is to train the PointMultiPlaneNeRF architecture
\begin{equation}\label{eq:multiplanenerffewshot}
\begin{array}{l}
    \F_{\text{PointMultiPlaneNeRF}}(\x , {\bf z}, {\bf d},\mathcal{S}; \theta)
    = ( {\bf c} , \sigma),
\end{array}
\end{equation}
which we treat as a target network modeling universal weights.

The main idea behind our \our{} approach is to use information extracted from selected support data (instead of many gradient descent iterations) to find an optimal task-related parameter update. This allows the \our{} target network weights to be switched between completely different 3D objects.

Specifically, we distinguish in a hypernetwork architecture $\mathcal{H}(\cdot;\delta)$ a trainable encoder, 
which transforms the ImagePlanes $I_1, \ldots, I_k$ (where $k\leq n$ is treated as a hyperparameter), hereafter called HyperPlanes, together with the corresponding viewing directions ${\bf d}_{I_1}, \ldots, {\bf d}_{I_k}$ and the current weights $\theta$ of the target network given by Eq.~\eqref{eq:multiplanenerffewshot}, into their low-dimensional code. The resulting representation is then propagated to produce the required update $\Delta \theta$. Thus, the new weights 
$\theta_{\mathcal{T}}$ are calculated using the following formula:
\begin{equation}\label{eq:paramupdate}
 \theta_{\mathcal{T}} = \theta_{\mathcal{T}}(\theta,\delta)=\theta  + \Delta \theta,
\end{equation}
where
\begin{equation}\label{eq:paramupdate1}
 \Delta\theta = \mathcal{H}(I_1, \ldots, I_k,{\bf d}_{I_1}, \ldots, {\bf d}_{I_k},\theta;\delta).
\end{equation}

The data and architecture of the \our{} model are visualized in Figure~\ref{fig:schema}). In the following paragraph, we present the successive steps of the training procedure.

\paragraph{Training procedure}
Training the \our{} model is done in the following steps. First, we sample a batch $\mathcal{B}$
from the given dataset $\mathcal{D}$ of training tasks such as in Eq.~\eqref{eq:nerftask}. Next, for each task $\mathcal{T}\in \mathcal{B}$ we compute new task-specific weights $\theta_{\mathcal{T}}$ according to Eqs.~\eqref{eq:paramupdate} and \eqref{eq:paramupdate1}.
Finally, we update $\theta$ (global target network weights) and $\delta$ (hypernetwork weights) in separate gradient steps using the Adam optimizer \cite{kingma2014adam} and the following loss function:
\begin{equation}\label{eq:HyperPlanesLoss}
\begin{array}{l}
{\mathcal L}_{\text{\our{}}}(\cdot;\theta,\delta)\\  \ \ \ =\sum_{\mathcal{T} \in \mathcal{B} ,J_i\in \mathcal{Q}}
\L_\text{PointMultiPlaneNeRF}(\cdot,{\bf d}_{J_i},\mathcal{S};\theta_{\mathcal{T}}),
\end{array}
\end{equation}
where ${\bf d}_{J_i}$ is the viewing direction corresponding to the query ImagePlane $J_i$.

For clarity, the entire training procedure is provided in Algorithm \ref{algorithm:HyperMAML} (see the appendix).

\begin{figure}[ht]
    \centering
  \textbf{\small  Ground truth} \quad \ \ \ \textbf{\small MultiPlaneNeRF} \ \ \ \ \ \textbf{\small \our{}}   \\
    \includegraphics[width=0.15\textwidth, trim=10 60 10 20, clip]{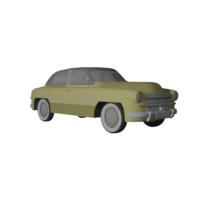}
    \includegraphics[width=0.15\textwidth, trim=10 60 10 20, clip]{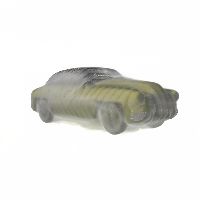}  
    \includegraphics[width=0.15\textwidth, trim=10 60 10 20, clip]{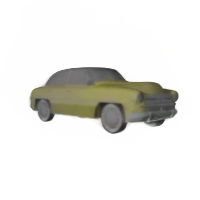}  \\
    \includegraphics[width=0.15\textwidth, trim=10 40 10 40, clip]{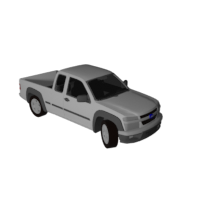}
    \includegraphics[width=0.15\textwidth, trim=10 40 10 40, clip]{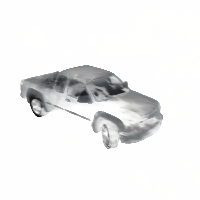}  
    \includegraphics[width=0.15\textwidth, trim=10 40 10 40, clip]{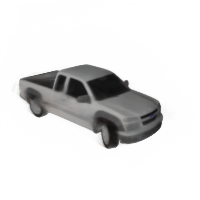}  \\
    \includegraphics[width=0.12\textwidth, trim=10 10 10 10, clip]{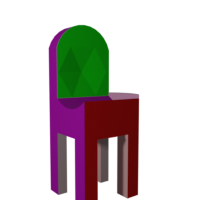}
    \quad
    \includegraphics[width=0.12\textwidth, trim=10 10 10 10, clip]{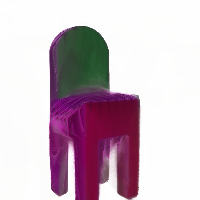}  
    \quad
    \includegraphics[width=0.12\textwidth, trim=10 10 10 10, clip]{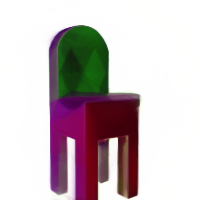}  \\
    \includegraphics[width=0.12\textwidth, trim=10 10 10 10, clip]{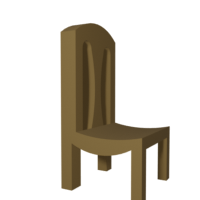}
    \quad
    \includegraphics[width=0.12\textwidth, trim=10 10 10 10, clip]{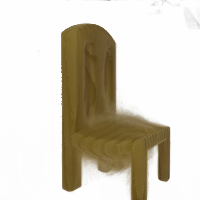}  
    \quad
    \includegraphics[width=0.12\textwidth, trim=10 10 10 10, clip]{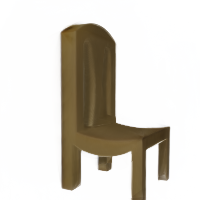}  \\ 
    \includegraphics[width=0.14\textwidth, trim=10 40 10 40, clip]{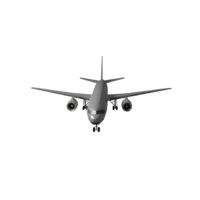}
    \includegraphics[width=0.14\textwidth, trim=10 40 10 40, clip]{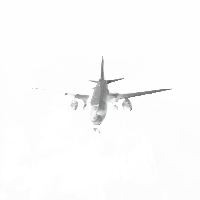}  
    \includegraphics[width=0.14\textwidth, trim=10 40 10 10, clip]{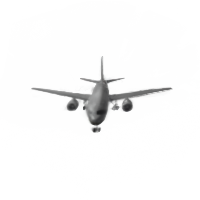}  \\
    \quad
    \includegraphics[width=0.10\textwidth, trim=0 10 10 30, clip]{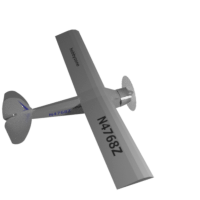}
    \qquad
    \includegraphics[width=0.10\textwidth, trim=0 10 10 30, clip]{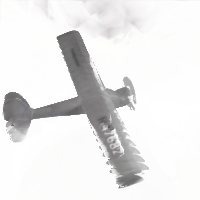}  
    \qquad
    \includegraphics[width=0.10\textwidth, trim=0 10 10 30, clip]{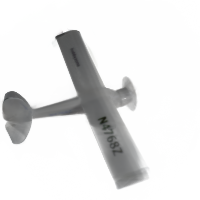}      
    \caption{Sample images from the MultiPlaneNeRF and HyperPlanes-100 models trained on the ShapeNet 200$\times$200 dataset, showing reconstructed Cars, Chairs, and Planes alongside their ground truths. All images were taken from the same viewing direction. For more examples, see Figure \ref{fig:reconstruction_big} in the appendix.}
    \label{fig:reconstruction}
\end{figure}


\paragraph{Fine-tuning} 
During inference in a few-shot learning scenario, when both sets of support and query samples are available, we can pre-train the hypernetwork on the support set while keeping the target network frozen. This approach refines the hypernetwork update for a specific object, yielding notably improved results with just a few fine-tuning iterations. Importantly, since the fine-tuning is exclusively performed on the support set, there is no risk of data leakage.

\section{Experiments}


In this section, we perform an experimental analysis of our approach. In Section \ref{sec:shapenet}, we thoroughly evaluate the \our{} model on the ShapeNet dataset~\cite{chang2015shapenet}. Section \ref{sec:ablation} details an ablation study of different components of the proposed solution. As performance evaluation measures, we employ peak signal-to-noise ratio (PSNR) and structural similarity index (SSIM) (see, e.g.,  \cite{wang2004image}). For a qualitative evaluation, see Figure \ref{fig:reconstruction}. Our source code is available at
\url{https://github.com/gmum/HyperPlanes}.



\subsection{Evaluation on the ShapeNet dataset}
\label{sec:shapenet}

\paragraph{Hypernetwork vs. gradient-based update}
In Table ~\ref{tab:shapenet}, we compare \our{} with a set of models proposed in \cite{tancik2021learned} that were trained with the REPTILE algorithm~\cite{nichol2018first}, which requires gradient-based fine-tuning to quickly adapt to the representation of an unseen object at hand. Each of the reference approaches requires 2000 fine-tuning steps. The comparison is made on the car, chair, and lamp classes of the ShapeNet 128$\times$128 dataset, and in each case we use $5$ HyperPlanes as input to the hypernetwork. The results are reported in terms of PSNR values between the true and rendered images. It is clear that our \our{} model achieves superior results, even though it does not require gradient-based fine-tuning to adapt to the object at hand.




\begin{table}[ht]
\centering
\caption{Comparison of HyperPlanes with the few-shot learning methods proposed in \cite{tancik2021learned} in terms of the PSNR ($\uparrow$) metric. All models were trained on the car, chair, and lamp classes of ShapeNet 128$\times$128. It is evident that our solution provides superior results for different object categories.}
\begin{tabular}{cccc}
\toprule
                                & Chairs                       & Cars                         & Lamps                        \\
\midrule
Standard                       & 12.49                        & 11.45                        & 15.47                        \\
MV Matched                     & 16.40                        & 22.39                        & 20.79                        \\
MV Shuffled                    & 10.76                        & 11.30                        & 13.88                        \\
MV Meta                        & 18.85                        & 22.80                        & 22.35                        \\
SV Meta                        & 16.54                        & 22.10                        & 20.95                        \\           
\our{}  (ours) & \bf 24.34 & \bf 27.80                       & \bf 25.95 \\
\bottomrule
\end{tabular}
\label{tab:shapenet}
\end{table}

\paragraph{\our{} fine-tuning}
In Table~\ref{tab:cross}, we show that \our{} benefits from additional gradient-based fine-tuning. We compare MultiPlaneNERF with \our{} (using 25 HyperPlanes) trained on the ShapeNet 200$\times$200 car, chair, and plane classes. During the inference, we fine-tune the hypernetwork component of the model for 100 steps to the object at hand. The results show that \our{} significantly outperforms MultiPlaneNeRF.


\paragraph{Generalization between object types}
In Table \ref{tab:cross}, we demonstrate the versatility of our solution across different object types. We compare MultiPlaneNeRF and \our{} on the task of rendering objects from the classes on which the models were not trained. We also analyze the change in PSNR with different numbers of fine-tuning iterations. The results show that \our{} with 100 inference steps outperforms the standard MultiPlaneNeRF in most cases, \our{} without fine-tuning, and \our{} with only 10 fine-tuning steps. Notably, even with only 10 fine-tuning steps, our model shows significant performance gains over MultiPlaneNeRF in most experimental settings.


\begin{table*}[ht]
\centering
\caption{The PSNR ($\uparrow$) values between ground truth and reconstructed images obtained by MultiPlaneNeRF, HyperPlanes without fine-tuning, HyperPlanes with 10 fine-tuning iterations (HyperPlanes-10), and HyperPlanes with 100 fine-tuning iterations (HyperPlanes-100) trained and tested on different classes (car, chair, and plane) of the ShapeNet 200$\times$200 dataset. All results were averaged over 5 runs.
}
{\small
\begin{tabular}{ccccccc}
\toprule
Trained on              & Tested on & MultiPlaneNeRF & HyperPlanes-0   & HyperPlanes-10  & HyperPlanes-100  & \\ \midrule
\multirow{3}{*}{Cars}   & Cars        & 24.91           & 28.48               & 29.31                 & \textbf{29.89}                                       \\
                        & Chairs      & 22.15           & 18.58               & 20.21                  & \textbf{23.13}                                       \\
                        & Planes      & 21.32           & 20.26               & 
21.58                 & \textbf{25.03}                                     \\ \midrule
\multirow{3}{*}{Chairs} & Cars        & 24.41           & 24.64              & 26.14                 & \textbf{27.57}                                     \\
                        & Chairs      & 24.26           & 22.95              & 25.07                 & \textbf{27.08}                                   \\
                        & Planes      & 20.84           & 18.36              & 21.49                 & \textbf{26.03}                                    \\ \midrule
\multirow{3}{*}{Planes} & Cars        & 24.19           & 13.57              & 22.35              & \textbf{25.35}                                   \\
                        & Chairs      & \textbf{21.69}           & 13.15              & 16.77                 & 
21.12                                  \\
                        & Planes      & 24.27           & 26.94              & 28.08                 & \textbf{28.58}                                 \\ \bottomrule
                        
\end{tabular}
}
\label{tab:cross}
\end{table*}

\paragraph{HyperPlanes inference vs. NeRF training}
We compare the number of iterations required for the Vanila NeRF to converge to a similar level of PSNR as a single HyperPlanes update (without fine-tuning) when presented with an unseen object. Specifically, we selected two objects from the car and chair classes of ShapeNet 128$\times$128, represented by 50 ImagePlanes taken from different perspectives. We divide the set of ImagePlanes into 25 training and 25 test images for NeRF training and, in the case of \our{}, use the training images as a support set to generate the necessary update. The results (averaged over 3 runs) shown in Figure \ref{fig:single_update_vs_nerf} highlight the advantageous generalization capabilities of \our{}, which achieves superior results in a single step compared to NeRF trained for approximately 36000 epochs.

\begin{figure}[ht]
    \centering
    \includegraphics[width=0.45\textwidth]{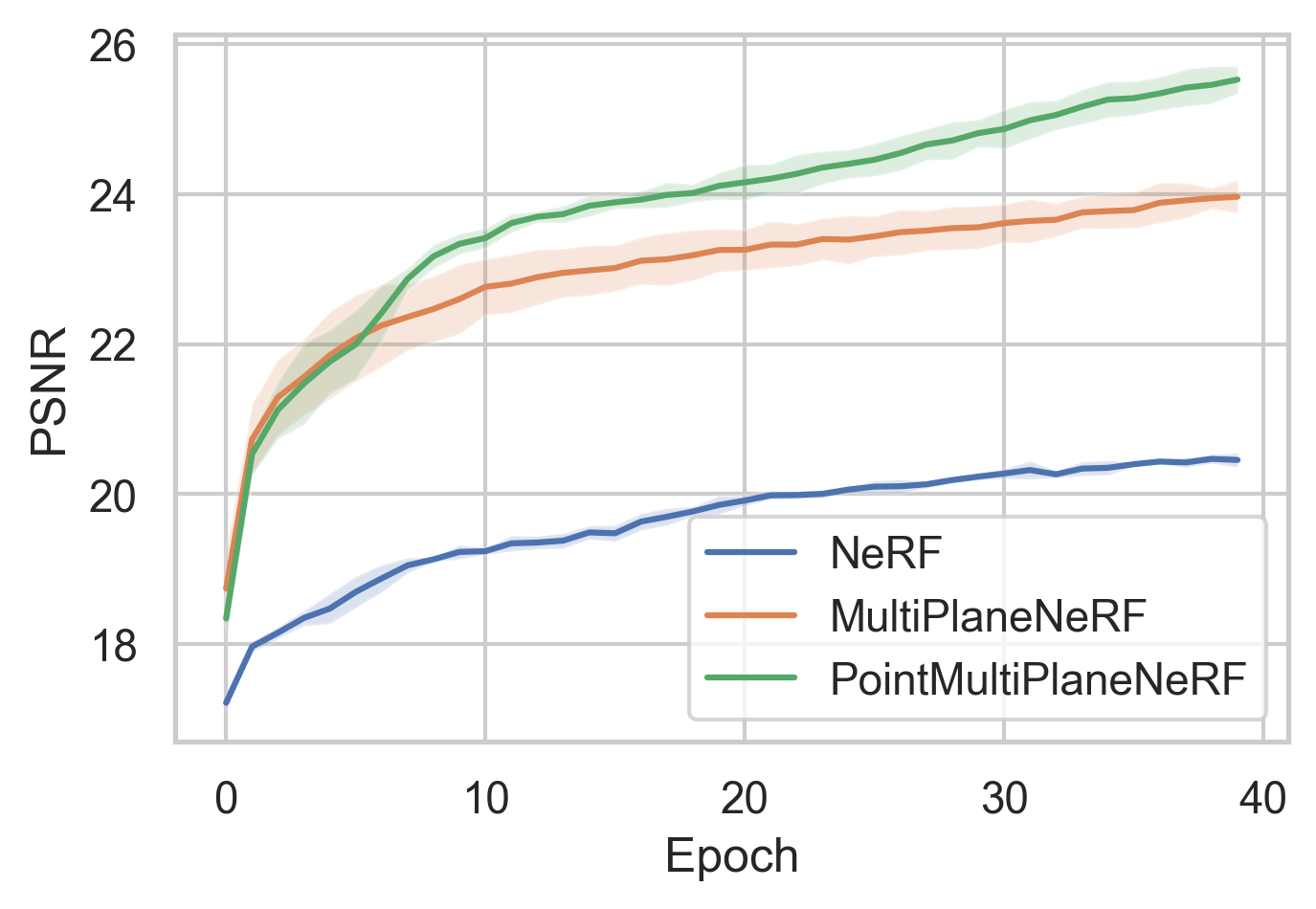}
    \caption{Comparison of PSNR ($\uparrow$) of the \our{} training process when using different target network architectures (NeRF, MultiPlaneNeRF, and PointMultiPlaneNeRF) and without weights and viewing directions in the hypernetwork input. 
 Results (averaged over 5 runs) were obtained on the car class of the ShapeNet 128$\times$128 dataset after 40 epochs. For extended results, see Figure \ref{fig:multiplane_vs_nerf_full} in the appendix.}
    \label{fig:multiplane_vs_nerf}
\end{figure}

\subsection{Ablation study} 
\label{sec:ablation}

\paragraph{Selection of a target network architecture}


We need to explain why PointMultiPlaneNeRF is the foundation of our target network instead of NeRF or MultiPlaneNeRF. As shown in Figure~\ref{fig:multiplane_vs_nerf}, it is clear that PointMultiPlaneNeRF learns much faster in the \our{} setting compared to MultiPlaneNeRF and NeRF. The benefits of including PointMultiPlaneNeRF as the target network extend beyond the training process, as shown in Table~\ref{tab:multiplane_vs_nerf}, which illustrates the differences observed on the test set measured by PSNR and SSIM.
 
\begin{table}[ht]
\centering
\caption{Comparison of the performance (in terms of PSNR ($\uparrow$)) of \our{} on the test dataset when using different target network architectures.
     Results (averaged over 5 runs) were obtained on the car class of the ShapeNet 128$\times$128 dataset after 40 epochs and are presented with their standard deviations.}
\resizebox{\columnwidth}{!}{%
\begin{tabular}{cccc}
\toprule
& NeRF & MultiPlaneNeRF & PointMultiPlaneNeRF \\ \midrule
PSNR & 19.78 $\pm$ 0.22 & 23.96 $\pm$ 0.19  & \textbf{24.35} $\pm$ 0.27 \\
SSIM & 0.881 $\pm$ 0.002 & 0.912 $\pm$ 0.001 & \textbf{0.927} $\pm$ 0.002 \\
\bottomrule
\end{tabular}
}
\label{tab:multiplane_vs_nerf}
\end{table}

\begin{figure}[ht]
\centering
\includegraphics[width=0.45\textwidth]{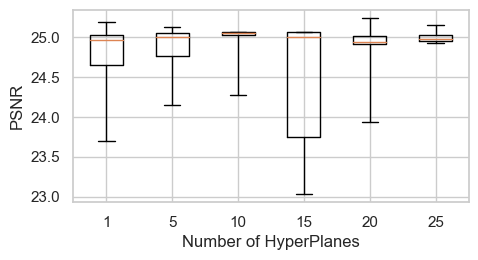}
\caption{Boxplots for PSNR ($\uparrow$) final values obtained by the HyperPlanes model with different numbers of HyperPlanes. Results were obtained on the planes class of the ShapeNet 200$\times$200 dataset after 40 epochs. For extended results, see Figure \ref{fig:hyperplanes_full} in the appendix.}
\label{fig:number_of_hyperplanes}
\end{figure}

\paragraph{Selection of the number of HyperPlanes}
We analyze the influence of the number of HyperPlanes used as input to the hypernetwork. As shown in Figure \ref{fig:number_of_hyperplanes}, all considered setups yield comparable results on average. However, architectures that use a smaller number of HyperPlanes show distinct instability, leading to significant variations in results over multiple runs. In contrast, the \our{} model using 25 planes demonstrates remarkable stability, with results showing minimal variation between iterations.

\begin{figure}[ht]
    \centering
    \includegraphics[width=0.45\textwidth]{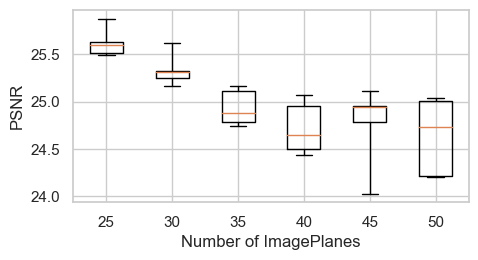} 
    \caption{Boxplots for PSNR ($\uparrow$) final values obtained by the HyperPlanes model with different numbers of ImagePlanes. Results were obtained on the planes class of the ShapeNet 128$\times$128 dataset after 40 epochs. For extended results, see Figure \ref{fig:imageplanes_full} in the appendix.
    }
    \label{fig:imageplanes}
\end{figure}

\paragraph{Selection of the number of ImagePlanes}
The authors of \cite{zimny2023multiplanenerf} suggest using 45 ImagePlanes as input to MultiPlaneNeRF. However, this poses a challenge in our scenario because the ShapeNet dataset contains only 50 plane images of each 3D object, which need to be split between sets of support and query samples. Our default configuration assigns 25 support ImagePlanes and 25 query ImagePlanes. In Figure \ref{fig:imageplanes}, we present the results of an experiment in which we explore how the performance of our approach changes as we add more and more (from 25 to 50) support ImagePlanes to the target network (we treat the sets of support and query samples as unified). It is evident that using 25 support ImagePlanes is an optimal choice, implying that including additional planes introduces unnecessary noise to our solution.

\begin{table}[ht]
\centering
\caption{Time (in seconds) required to infer an image from a single viewing direction, obtained on the car class of ShapeNet 200$\times$200. Results (averaged over all samples) are presented with their standard deviations.}
\begin{tabular}{ccc}
\midrule
HyperPlanes-0 & HyperPlanes-10 & HyperPlanes-100 \\ \midrule
25.41 $\pm$ 0.14      &  27.13 $\pm$ 0.12          & 41.3 $\pm$ 0.14    \\ \bottomrule
\end{tabular}
\label{tab:time}
\end{table}
\vspace{-0.5cm}

\begin{figure}[ht]
    \centering
    
    \begin{subfigure}{0.45\textwidth}  
        \includegraphics[width=\linewidth]{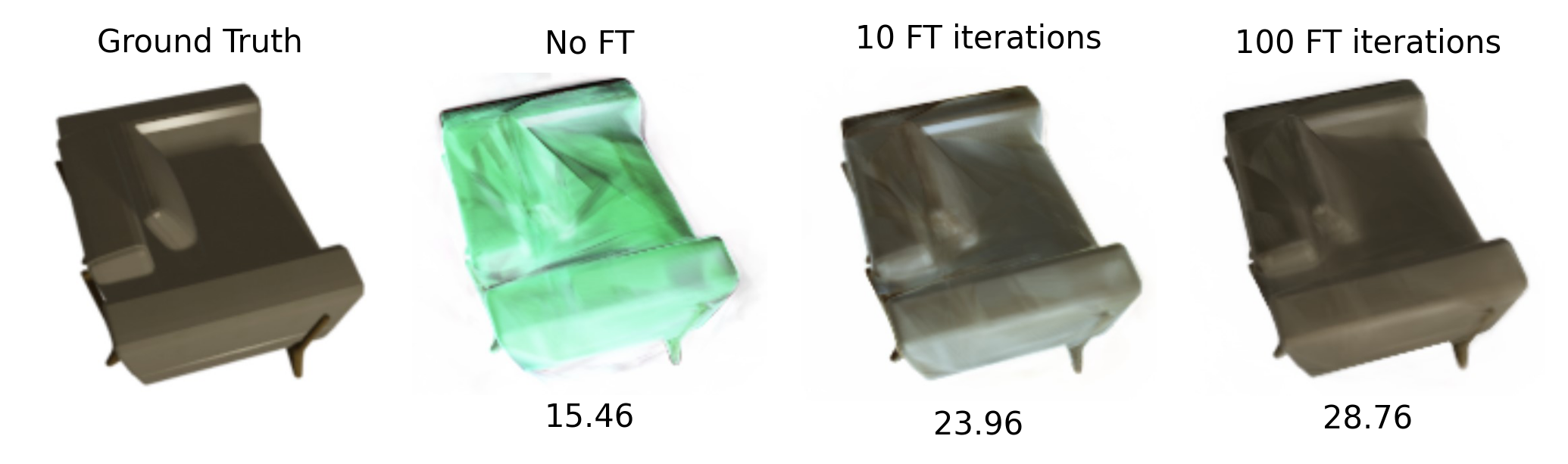}
        \caption{Sample single-view images inferred after training on the chair class of ShapeNet 200$\times$200.}
    \end{subfigure}
    
    \vspace{1em}  
    
    \begin{subfigure}{0.45\textwidth}  
        \includegraphics[width=\linewidth]{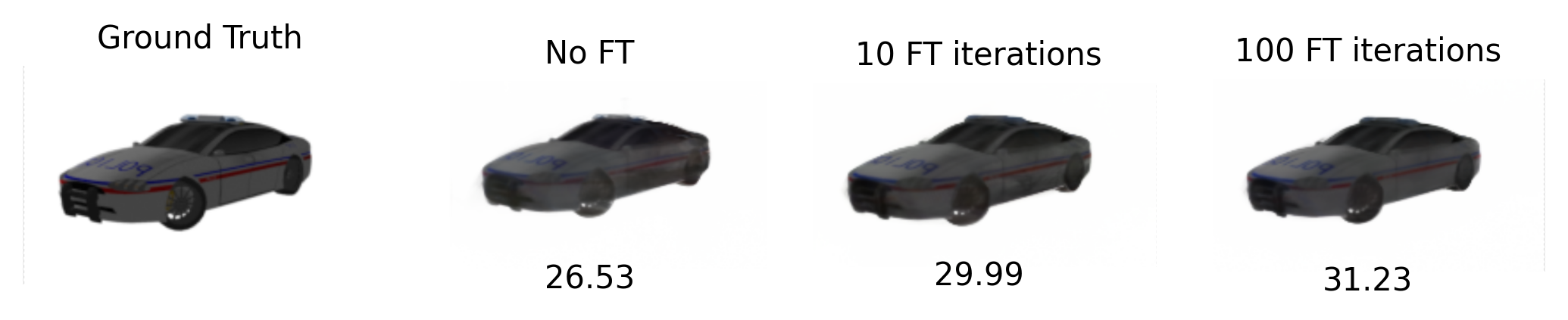}
        \caption{Sample single-view images inferred after training on the car class of ShapeNet 200$\times$200.}
    \end{subfigure}
    
    \caption{Comparison of the inference capabilities of the HyperPlanes model with respect to the number of fine-tuning (FT) iterations. The PSNR ($\uparrow$) values shown were calculated for entire 3D objects.}
    \label{fig:fine_tuning}
\end{figure}

\paragraph{Fine-tuning}
We perform an experiment that emphasizes the importance of using fine-tuning techniques during inference. Despite the high performance of the trained \our{} model, some examples in the test set show suboptimal reconstruction, as shown in Figure \ref{fig:fine_tuning}. In contrast, only 10 iterations of pre-training (using support samples) of the hypernetwork (with the target network frozen) yield significantly improved results. In addition, the results are even better after 100 iterations of fine-tuning. 
In the Table \ref{tab:time} we present the time differences between various inference setups. Intuitively, performing additional fine-tuning iterations requires more processing time. This introduces a trade-off between inference time and the quality of the desired results. Achieving optimal outcomes involves navigating this balance between training time and result quality.

\paragraph{Impact of the content of the \our{} model architecture}

\begin{figure}[ht!]
    \centering
    \includegraphics[width=0.42\textwidth]{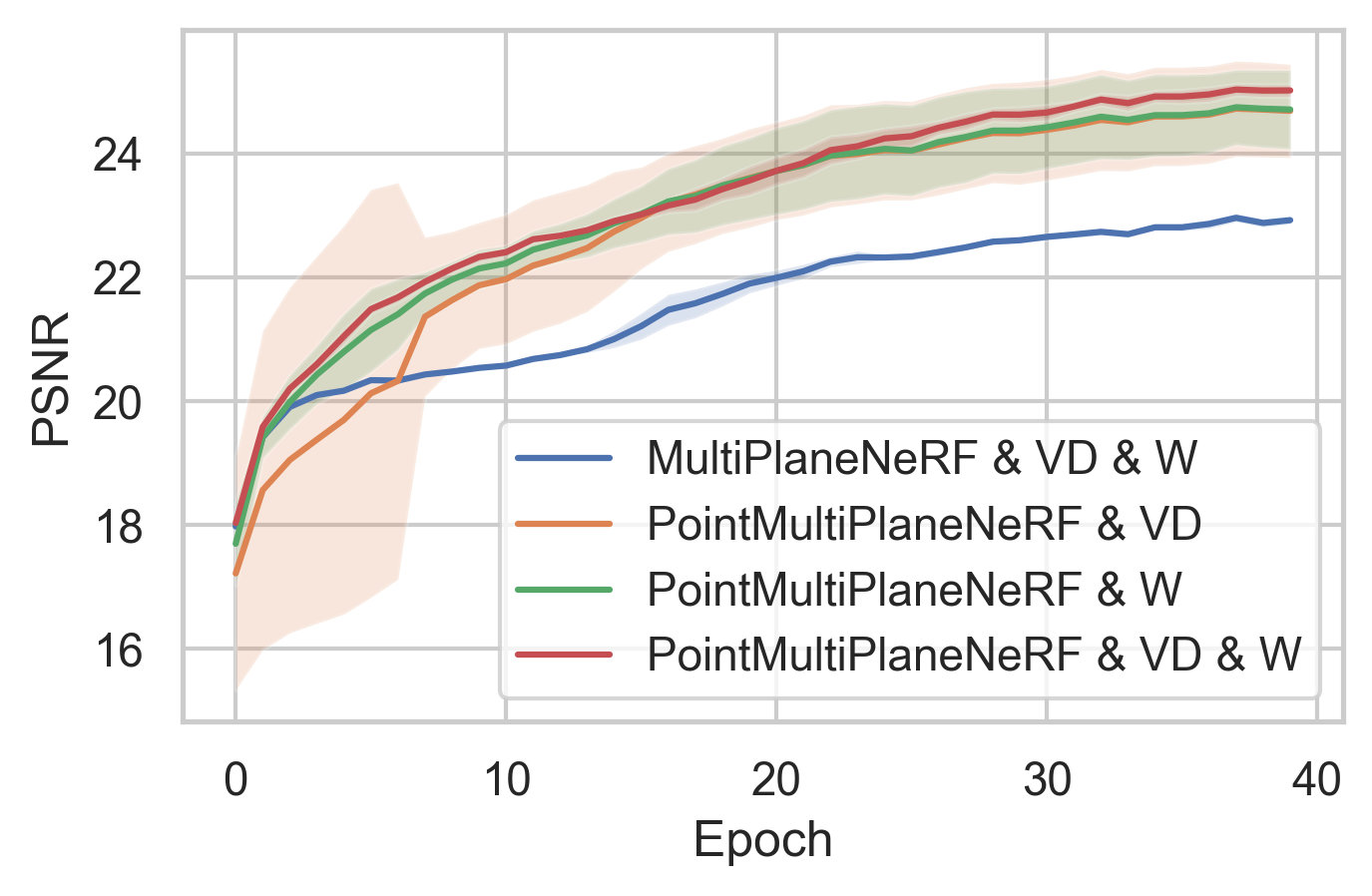}
    \caption{Impact of using weights (W) and view directions (VD) as hypernetwork input, and MultiPlaneNeRF instead of PointMultiPlaneNeRF as target network, expressed in terms of PSNR ($\uparrow$). Results (averaged over 5 runs) were obtained on the planes of the ShapeNet 200$\times$200 dataset after 40 epochs. For extended results, see Figure \ref{fig:hypernetwork_comparison_full} in the appendix.}
    \label{fig:hypernetwork_comparison}
    \vspace{-0.5cm}
\end{figure}

We explore the importance of including different input information in the architecture of \our{}, i.e., the weights and view directions in the hypernetwork input, as well as the coordinates of the rendered point in the target network (which distinguishes our PointMultiPlaneNeRF from MultiPlaneNeRF).
The results of a conducted experiment, shown in Figure \ref{fig:hypernetwork_comparison}, demonstrate a remarkable discrepancy in PSNR between MultiPlaneNeRF and PointMultiPlaneNeRF backbone. 
While excluding weights or view directions from the hypernetwork input results in a smaller difference, it contributes significantly to the stabilization of the training process, ensuring consistent results over multiple runs.

\section{Conclusions}
The paper presents a few-shot training approach for NeRF representation of 3D objects using the hypernetwork paradigm. Our solution does not require gradient optimization during inference, resulting in the novel  \our{} model, which is a computationally efficient single-step inference method that requires only a small number of input images.
This leads to an exceptionally accelerated object reconstruction (even 380 times faster than the vanilla NeRF trained for 36000 epochs), which was confirmed by direct experimental comparison with existing state-of-the-art solutions and an extensive ablation study.
 As a result, we have developed a technique that enables the instant generation of views of new, unseen 3D objects, making it applicable to spatial computing and other computer graphics applications.


\paragraph{Limitations}
A potential drawback of the proposed solution is that it may not be able to generate reconstructions of comparable quality to those generated by a vanilla NeRF with extremely extensive training time. This issue will be addressed as a starting point for our future work. 
\paragraph{Societal impact}
Our research aims to advance the field of machine learning. It has many potential societal impacts, none of which we feel need to be highlighted here.



\appendix
\onecolumn
\newpage
\section{Additional experimental study}
\begin{figure}[ht]
    \centering
    \includegraphics[width=0.98\textwidth]{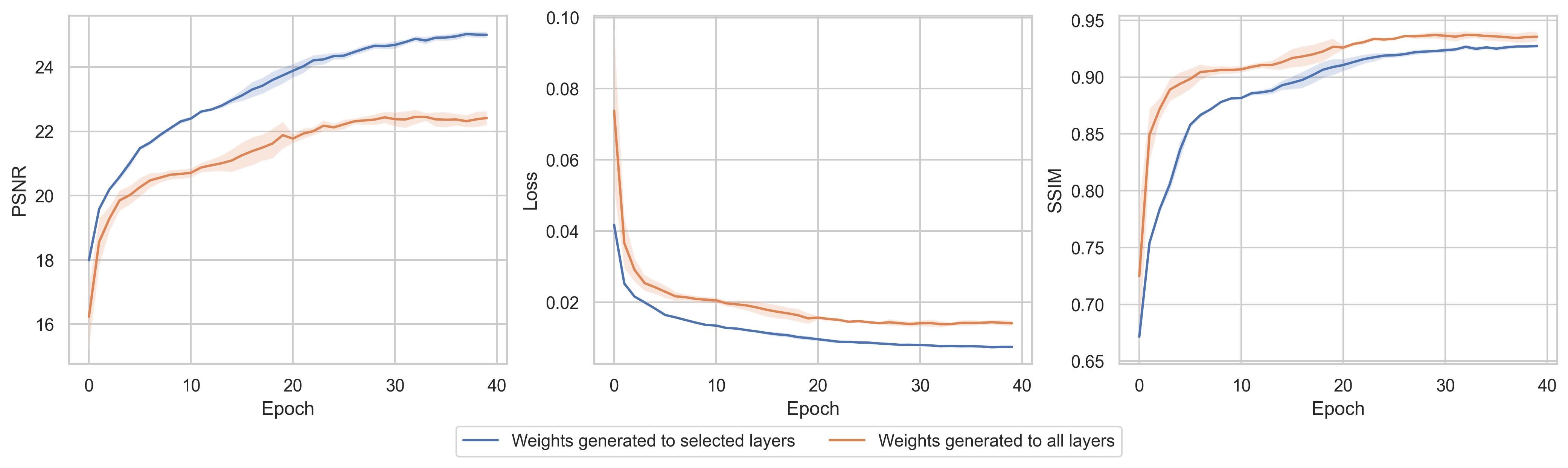} 
    \caption{Comparison of the \our{} training process when generating weights for all layers versus selected layers, using PSNR ($\uparrow$), loss ($\downarrow$), and SSIM ($\uparrow$). Results (averaged over 5 runs) were obtained on the plane class of the ShapeNet 200$\times$200 dataset after 40 epochs.
     The loss was multiplied by 100.}
    \label{fig:hypernetwork_generate_full}
\end{figure}

\paragraph{Selection of updated target network layers}
This experiment aims to investigate the reasoning behind selectively generating updates for specific layers in the target network. Figure \ref{fig:hypernetwork_generate_full} examines two scenarios: one in which updates are generated for all target network parameters, and the other in which updates are generated only for the last layer, i.e., the RGB color and opacity layer. The experiment demonstrates a significant performance advantage for the scenario where updates are generated only for the last selected layers. A possible explanation for this phenomenon is that generating updates for only a few layers is essential, while generating updates for more layers introduces unnecessary noise that the hypernetwork cannot handle effectively.

\begin{figure}[ht]
    \centering
    \includegraphics[width=0.98\textwidth]{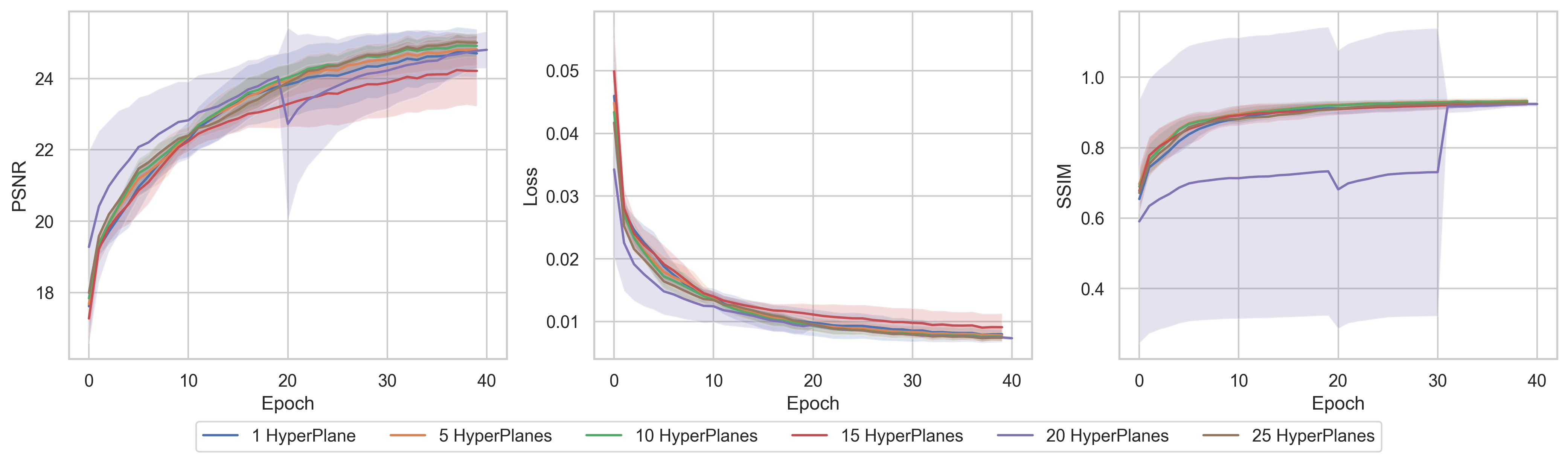} 
    \caption{Comparison of the \our{} training process when different numbers of HyperPlanes are provided to the hypernetwork, using PSNR ($\uparrow$), loss ($\downarrow$), and SSIM ($\uparrow$). Results (averaged over 5 runs) were obtained on the plane class of the ShapeNet 200$\times$200 dataset after 40 epochs.
     The loss was multiplied by 100.}
    \label{fig:hyperplanes_full}
\end{figure}

\paragraph{Selection of the number of HyperPlanes}
Figure \ref{fig:hyperplanes_full} displays the results of our experiment on the impact of selecting the number of HyperPlanes. It is clear that using 25 HyperPlanes ensures stable outcomes.

\begin{figure}[ht]
    \centering
    \includegraphics[width=0.98\textwidth]{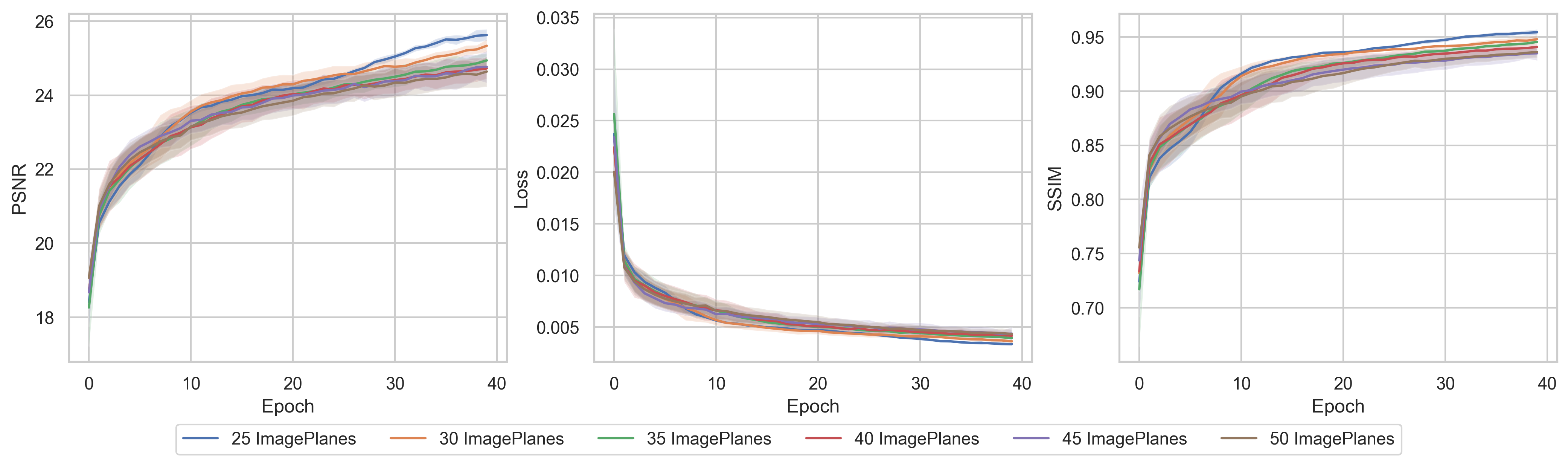} 
    \caption{Comparison of the \our{} training process when different numbers of ImagePlanes are provided to the target network, using PSNR ($\uparrow$), loss ($\downarrow$), and SSIM ($\uparrow$). Results (averaged over 5 runs) were obtained on the plane class of the ShapeNet 200$\times$200 dataset after 40 epochs.
     The loss was multiplied by 100.}
    \label{fig:imageplanes_full}
\end{figure}

\paragraph{Selection of the number of ImagePlanes}
Figure \ref{fig:imageplanes_full} shows the results of our experiment on the impact of selecting the number of ImagePlanes. The results indicate that utilizing 25 ImagePlanes has clear benefits.

\begin{figure}[ht]
    \centering
    \includegraphics[width=0.98\textwidth]{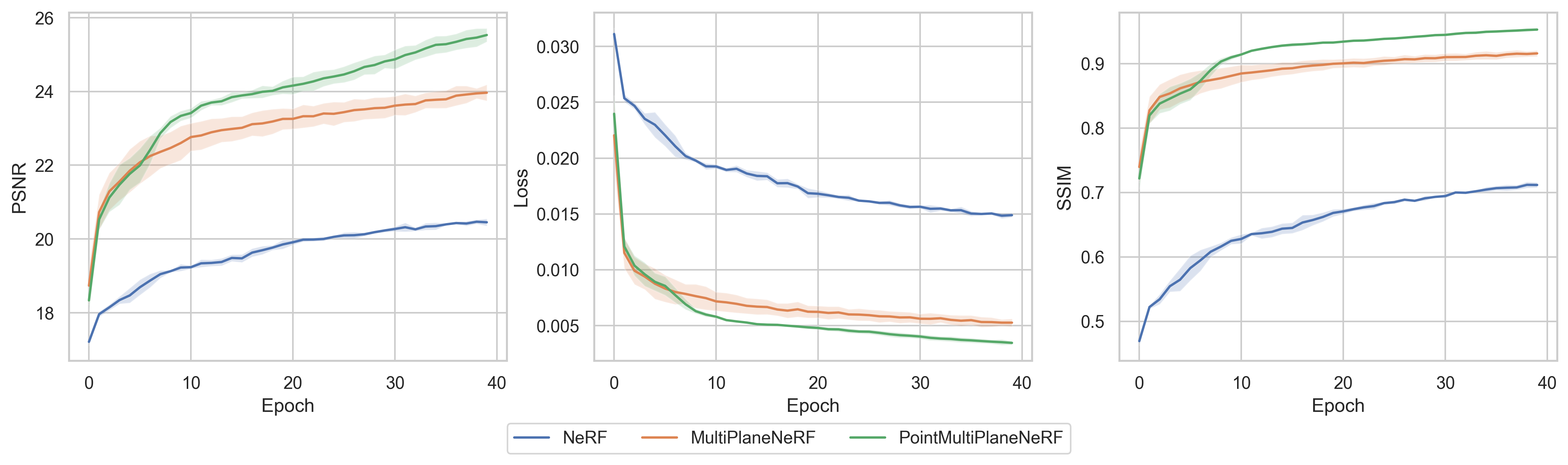} 
    \caption{Comparison of the \our{} training process when different target network architectures are employed, using PSNR ($\uparrow$), loss ($\downarrow$), and SSIM ($\uparrow$). 
     Results (averaged over 5 runs) were obtained on the car class of the ShapeNet 128$\times$128 dataset after 40 epochs.
      The loss was multiplied by 100.}
    \label{fig:multiplane_vs_nerf_full}
\end{figure}

\paragraph{Selection of a target network architecture} 
Figure \ref{fig:multiplane_vs_nerf_full} shows the results of our experiment where we tested different target networks as the backbone for the HyperPlanes model. The findings clearly demonstrate the benefits of using the PointMultiPlaneNeRF architecture.

\begin{figure}[ht]
    \centering
    \includegraphics[width=0.98\textwidth]{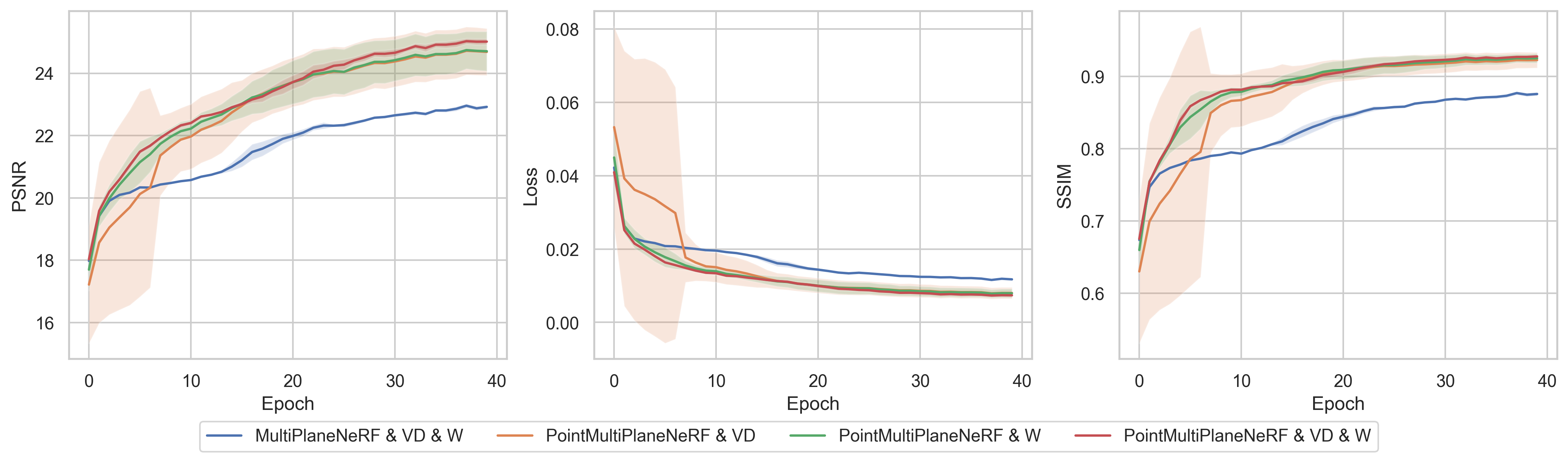}
    \caption{Impact of using weights (W) and view directions (VD) as hypernetwork input, and MultiPlaneNeRF instead of PointMultiPlaneNeRF as target network, expressed in terms of PSNR ($\uparrow$), loss ($\downarrow$), and SSIM ($\uparrow$). Results (averaged over 5 runs) were obtained on the car class of the ShapeNet 128$\times$128 dataset after 40 epochs. The loss was multiplied by 100.}
    \label{fig:hypernetwork_comparison_full}
\end{figure}

\paragraph{Impact of the content of \our{} architecture} 
The outcomes of our experiment on the impact of the content of the \our{} model architecture are presented in Figure \ref{fig:hypernetwork_comparison_full}. The results indicate the advantages of utilizing PointMultiPlaneNeRF as the target network, with viewing directions and weight values as the input for the hypernetwork.

\paragraph{Single-class inference capabilities} 
Table \ref{tab:single_inference} explores the capabilities of \our{} and MultiPlaneNeRF in single-class inference, where the models are trained and tested on the same class. The models were trained and tested on the same class. Our observations show that 10 fine-tuning iterations for the \our{} model produce significantly improved results compared to MultiPlaneNeRF.

\section{Training details}
\label{app:training-details}

\paragraph{Data} We used the ShapeNet dataset in two different configurations. In the first, we used the original resolution, following the conventional training and testing splits (ShapeNet 128$\times$128). In the second, we introduced a new configuration with higher resolution, accompanied by a revised training and testing split, analogous to that used in AtlasNet (ShapeNet 200$\times$200).

\paragraph{Architecture} Our architecture consists of three main components: hypernetwork, encoder (a part of the hypernetwork), and target network. Depending on the type of dataset, we use different parameter configurations for the hypernetwork and encoder to achieve better results. For example, for a smaller dataset, we used a ConvNet architecture as the encoder, while for larger datasets, we used ResNet101.

\paragraph{Hyperparameters} In all experiments, we used a batch size that contained only one task. For each task in the datasets, we chose the number of epochs. For different classes of Shapenet 128$\times$128 these were: Cars -- 1000, Lamps -- 1100, Chairs -- 700, while for different classes of Shapenet 200$\times$200 these were: Cars -- 2420, Planes -- 2460, Chairs -- 1920. Additionally, a grid search was performed in which we tested different learning rates (from 0.01 to 0.00000001), different numbers of HyperPlanes and ImagePlanes, and different configurations of the target network, such as number of layers and network width. For the ablation study, we run each experiment 5 times to obtain the mean and standard deviation estimates.

\paragraph{Implementation} 
We implemented \our{} in the PyTorch framework~\cite{paszke2019pytorch}. 
Training was performed on a single NVidia A100 GPU. 

\begin{figure*}[ht]
    \centering
    \includegraphics[width=\textwidth]{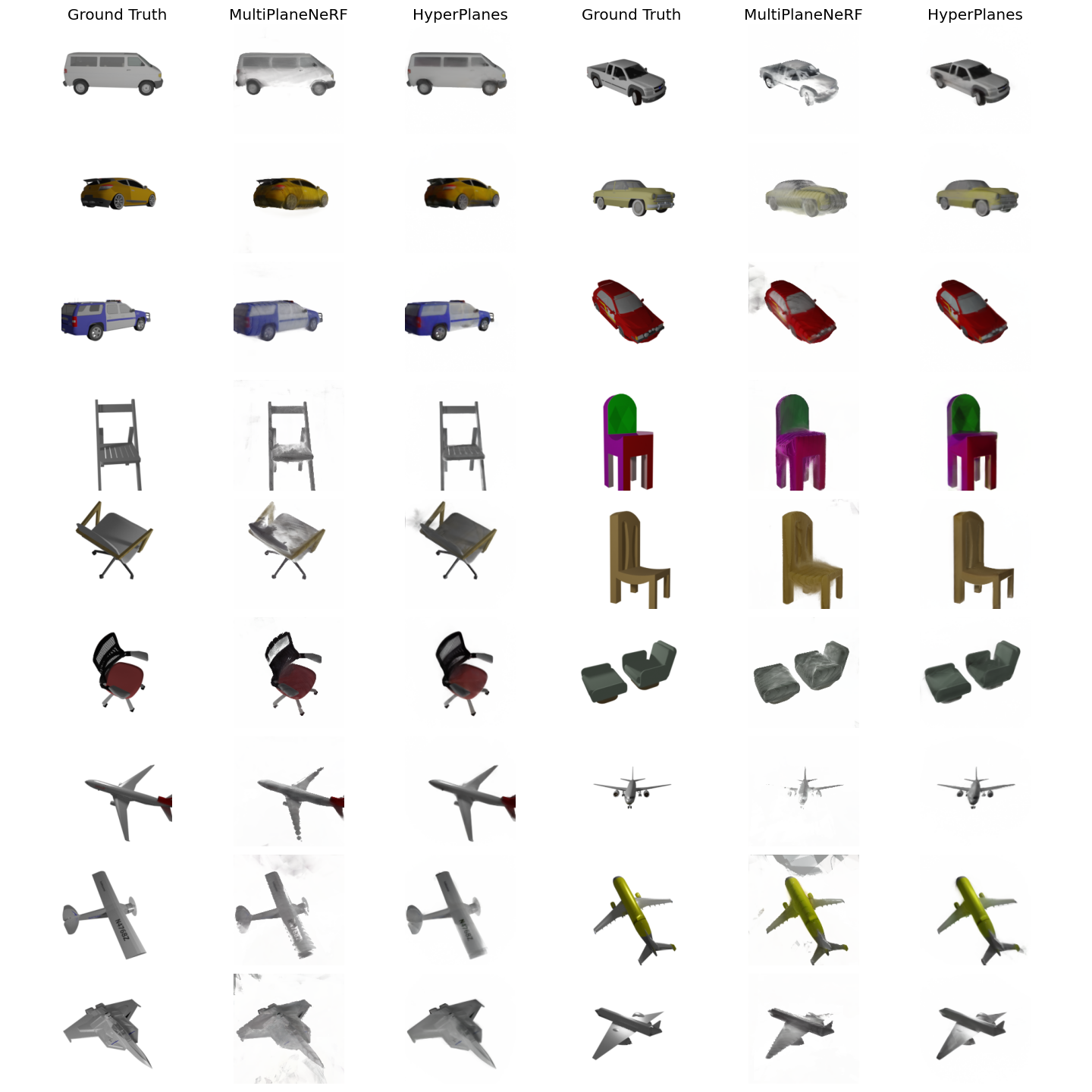} 
    \caption{Sample images from the MultiPlaneNeRF and HyperPlanes-100 models trained on the ShapeNet 200$\times$200 dataset, showing reconstructed cars, chairs, and planes alongside their ground truths. All images were taken from the same viewing direction.}
    \label{fig:reconstruction_big}
\end{figure*}

\begin{table*}[ht]
\centering
\caption{The PSNR ($\uparrow$) values on the train and test sets for MultiPlaneNeRF, Autoencoder, HyperPlanes without fine-tuning, HyperPlanes with 10 fine-tuning iterations (HyperPlanes-10), and HyperPlanes with 100 fine-tuning iterations (HyperPlanes-100).  Results (averaged over 5 runs) were obtained on the car, chair, and plane classes of the ShapeNet 200$\times$200 dataset.}
\begin{tabular}{ccccccc}

\toprule
                    &   & MultiPlaneNeRF & Autoencoder & HyperPlanes  & HyperPlanes-10  & HyperPlanes-100  \\ \midrule
\multirow{3}{*}{Train} & Cars   & 26.21          & 28.14       & \textbf{30.12}       & \textbf{30.12}        & \textbf{30.12}         \\
                       & Chairs & 25.28          & 23.90       & \textbf{28.30}       & \textbf{28.30}        & \textbf{28.30}         \\
                       & Planes & 25.28          & 24.83       & \textbf{30.12}       & \textbf{30.12}        & \textbf{30.12}         \\ \midrule
\multirow{3}{*}{Test}  & Cars   & 24.79          & 20.86       & 28.48                & 29.31                 & \textbf{29.89}         \\
                       & Chairs & 24.26          & 17.17       & 22.95                & 25.07                 & \textbf{27.08}         \\
                       & Planes & 24.26          & 14.18       & 26.94                & 28.08                 & \textbf{28.58}         \\ \bottomrule
\end{tabular}
\label{tab:single_inference}
\end{table*}

\begin{algorithm*}[tb]
   \caption{Training procedure for the \our{} model}
   \label{algorithm:HyperMAML}
\begin{algorithmic}[1]
   \STATE {\bfseries Require:} dataset of training tasks $\mathcal{D}$; hyperparameters $\beta$ (stepsize for the Adam optimizer), $k$ (number of HyperPlanes)
   \WHILE{ not done }
   \STATE sample batch of tasks $\mathcal{B}$ from $\mathcal{D}$
   \FOR{each task $\mathcal{T}=\{\mathcal{L}_{\text{PointMultiPlaneNeRF}}(\cdot,\mathcal{S};\theta),\mathcal{S}, \mathcal{Q}\}$  from  $\mathcal{B}$}
   \STATE compute new task-specific weights $ \theta_{\mathcal{T}}$ using Eqs. \eqref{eq:paramupdate} and \eqref{eq:paramupdate1}
   \ENDFOR
   \STATE update $\theta$ and $\delta$ using the Adam optimizer with the stepsize $\beta$ and the loss function $\L_{\text{\our{}}}(\cdot;\theta,\delta)$ given in Eq.~\eqref{eq:HyperPlanesLoss} 
    \ENDWHILE
\end{algorithmic}
\end{algorithm*}

\twocolumn

\paragraph{ShapeNet 128$\times$128}

\begin{itemize}
\setlength\itemsep{0.0em}
\item[]\textbf{Hypernetwork}
\begin{itemize}
\item[] fully connected layers with ReLU activation functions
\item[] layer depth: 3
\item[] layer width: 256
\item[] learning rate: 0.0001
\item[] ImagePlanes: 25
\item[] HyperPlanes: 5
\item[]\textbf{Encoder} (ConvNet)
\begin{itemize}
\item[] convolutional layers followed by batch normalization
\item[] kernel size 3$\times$3 for each layer
\item[] learning rate: 0.0001
\end{itemize}
\end{itemize}

\item[]\textbf{Target Network}
\begin{itemize}
\item[] fully connected layers with ReLU activation functions
\item[] layer depth: 8
\item[] layer width: 256
\item[] importance: 128
\item[] rand: 512
\item[] samples: 64
\item[] learning rate: 0.0001
\end{itemize}
\end{itemize}

\newpage
\paragraph{ShapeNet 200$\times$200}

\begin{itemize}
\setlength\itemsep{0.0em}
\item[]\textbf{Hypernetwork}
\begin{itemize}
\item[] fully connected layers with ReLU activation functions
\item[] layer depth: 10
\item[] layer width: 256
\item[] learning rate: 0.0001
\item[] ImagePlanes: 25
\item[] HyperPlanes: 25
\item[]\textbf{Encoder} (ResNet101)
\begin{itemize}
\item[] classic ResNet101 architecture with specific output size
\item[] learning rate: 0.0001
\end{itemize}
\end{itemize}
\item[]\textbf{Target Network} 
\begin{itemize}
\item[] fully connected layers with ReLU activation functions
\item[] layer depth: 8
\item[] layer width: 256
\item[] importance: 128
\item[] rand: 512
\item[] samples: 64
\item[] learning rate: 0.0001
\end{itemize}
\end{itemize}




\end{document}